\documentclass[12pt]{article}
\usepackage{algorithm}
\usepackage{algorithmic}
\usepackage{latexsym}
\usepackage{cite}
\usepackage{enumitem}
\usepackage{amssymb,mathrsfs}
\usepackage{booktabs}
\usepackage{subfigure}
\usepackage{float}
\usepackage{color}
\usepackage{amsmath}
\usepackage[colorlinks,
            linkcolor=blue,
            anchorcolor=red,
            citecolor=blue]{hyperref}
\usepackage{bm}
\usepackage{natbib}
\usepackage{graphicx}
\usepackage{float}
\usepackage{subfigure}
\usepackage{wrapfig}
\usepackage{amsthm}
\usepackage{epstopdf}
\usepackage{lscape}
\usepackage{multirow}
\usepackage[table,xcdraw]{xcolor}
\usepackage{booktabs}

\setcitestyle{round}
\allowdisplaybreaks[4]

\newtheorem{theorem}{Theorem}[part]

\def \TKFAC{\rm TKFAC}
\def \TEKFAC{\rm TEKFAC}

\begin{document}
\title{Eigenvalue-corrected Natural Gradient Based on a New Approximation 
}

\author {
        Kai-Xin Gao\textsuperscript{\rm 1}\thanks{Equal contribution} ,
        Xiao-Lei Liu\textsuperscript{\rm 1$\ast$},
        Zheng-Hai Huang\textsuperscript{\rm 1$\ast$},
        Min Wang\textsuperscript{\rm 2},\\
        Shuangling Wang\textsuperscript{\rm 2},
        Zidong Wang\textsuperscript{\rm 2},
        Dachuan Xu\textsuperscript{\rm 3}\thanks{Corresponding author, Email: xudc@bjut.edu.cn} ,
        Fan Yu\textsuperscript{\rm 2}\\
        {\small $^1$ School of Mathematics, Tianjin University \vspace{-0.5em}} \\
        {\small $^2$ Central Software Institute, Huawei Technologies \vspace{-0.5em}}\\
        {\small $^3$ Department of Operations Research and Information Engineering, Beijing University of Technology}
}

\date{}

\maketitle

\begin{abstract}
\noindent

\vspace{3mm}

 Using second-order optimization methods for training deep neural networks (DNNs) has attracted many researchers. A recently proposed method, Eigenvalue-corrected Kronecker Factorization (EKFAC) \citep{ekfac2018}, proposes an interpretation of viewing natural gradient update as a diagonal method, and corrects the inaccurate re-scaling factor in the Kronecker-factored eigenbasis. \citet{tkfac2020} considers a new approximation to the natural gradient, which approximates the Fisher information matrix (FIM) to a constant multiplied by the Kronecker product of two matrices and keeps the trace equal before and after the approximation. In this work, we combine the ideas of these two methods and propose Trace-restricted Eigenvalue-corrected Kronecker Factorization (TEKFAC). The proposed method not only corrects the inexact re-scaling factor under the Kronecker-factored eigenbasis, but also considers the new approximation method and the effective damping technique proposed in \citet{tkfac2020}. We also discuss the differences and relationships among the Kronecker-factored approximations. Empirically, our method outperforms SGD with momentum, Adam, EKFAC and TKFAC on several DNNs.

\vspace{3mm}


\end{abstract}

\section{Introduction}
Deep learning has made significant progress in various natural language and computer vision applications. But as models becoming more and more complex, deep neural networks (DNNs) usually have huge parameters (for example, VGG16 has over 1.5 million parameters) to be trained, which takes a long time. Therefore, the research of more efficient optimization algorithms has attracted many researchers.

Among the algorithms for training DNNs, the most popular and widely used method is Stochastic Gradient Descent (SGD) \citep{sgd}. During training, the goal of SGD is to find the optimal parameters $\omega$ to minimize the objective function $h(\omega)$. The parameters $\omega$ are updated by: $\omega\leftarrow\omega-\eta \nabla_{\omega}h$, where $\eta$ is the learning rate. To achieve better training performance, many variants of SGD also have been proposed, such as momentum \citep{sgdm}, Nesterov's acceleration \citep{nes1983} and etc. However, SGD only considers first-order gradient information,
which leads to some deficiencies, including relatively-slow convergence and sensitivity to hyper-parameter settings.

To avoid these problems, the second-order optimization algorithm may be a good choice. More importantly, second-order optimization algorithms can greatly accelerate convergence by using curvature matrix to correct gradient through training. The parameters update rule is: $\omega \gets \omega - \eta F^{-1} \nabla_\omega h$, where $F^{-1}$ is the inverse of curvature matrix. The curvature matrix $F$ is defined differently in second-order optimization algorithms. For Newton's method, $F$ is the Hessian matrix which represents second-order derivatives. For natural gradient method \citep{nat1998}, $F$ is the Fisher information matrix (FIM) which represents covariance of second-order gradient statistics. However, the curvature matrix and its inverse dramatically increase computing and storage costs. It is impractical to compute and invert an exact curvature matrix directly. Therefore, many approximate methods have been proposed.

A simple but crude method is diagonal approximation, such as AdaGrad \citep{ada2011}, RMSprop \citep{rms2012}, Adam \citep{ada2014} and etc. These algorithms are computationally tractable but lose much curvature matrix information. More elaborate algorithms are no longer limited to diagonal approximation. For Newton's methods, quasi-Newton method (\citeauthor{qn1977},\citeyear{qn1977}; \citeauthor{qn2011},\citeyear{qn2011}; \citeauthor{qn2019},\citeyear{qn2019}; \citeauthor{qn2020},\citeyear{qn2020}) can be used to approximate the Hessian matrix and its advantages over Newton's method is that the Hessian matrix does not need to be inverted directly. Hessian-Free optimization approach (\citeauthor{hf2010},\citeyear{hf2010}; \citeauthor{hf2013},\citeyear{hf2013}; \citeauthor{hf2017},\citeyear{hf2017}) provides a matrix-free conjugate-gradient algorithm for approximating the Hessian matrix. For natural gradient methods, Kronecker-factored Approximate Curvature (KFAC) \citep{kfac2015} presents efficient block diagonal approximation and block tri-diagonal approximation of the FIM in fully-connected neural networks. This method has been further extended to convolutional neural networks \citep{kfc2016}, recurrent neural networks \citep{kfacr2018} and variational Bayesian neural networks \citep{nkfac2018,nekfac2019}. KFAC has also been used in large-scale distributed computing for deep neural networks (\citeauthor{dis2017},\citeyear{dis2017}; \citeauthor{lar2019},\citeyear{lar2019}; \citeauthor{dis2020},\citeyear{dis2020}; \citeauthor{w2020},\citeyear{w2020}).

In particular, \citet{ekfac2018} proposes a new explanation for the natural gradient update, in which the natural gradient update is viewed as diagonal method in Kronecker-factored eigenbasis. And under this interpretation, the re-scaling factor under the KFAC eigenbasis is not exact. So Eigenvalue-corrected Kronecker Factorization (EKFAC) is proposed to correct the inaccurate re-scaling factor. Recently, \citet{tkfac2020} adopts a new model to approximate the FIM called Trace-restricted Kronecker Factorization (TKFAC). TKFAC approximates the FIM as a constant multiple of the Kronecker product of two matrices. In experiments, TKFAC has better performance than KFAC. Therefore, it is natural for us to consider the TKFAC's model under the interpretation proposed in EKFAC.

In this work, we combine the ideas of EKFAC and TKFAC, then present Trace-restricted Eigenvalue-corrected Kronecker Factorization (TEKFAC). Our contribution can be summarized as follows:
\begin{itemize}
  \setlength{\itemsep}{3pt}
  \setlength{\parsep}{0pt}
  \setlength{\parskip}{0pt}
  \item Instead of approximating FIM to the Kronecker product of two smaller matrices, we consider EKFAC based on the approximation model adopted by TKFAC. So, we change the Kronecker-factored eigenbasis in EKFAC and propose TEKFAC, which not only corrects the inexact re-scaling factor but also takes the advantages of TKFAC.
  \item We discuss the relationships and differences among the several methods, including KFAC, EKFAC, TKFAC and TEKFAC. Empirically, we compare TEKFAC with SGD with momentum (SGDM), Adam, EKFAC and TKFAC using the SVHN, CIFAR-10 and CIFAR-100 datasets on VGG16 and ResNet20. Our method has more excellent performance than these baselines.
\end{itemize}

\section{Methods to Approximate the Natural Gradient} \label{sec-2}
\subsection{Natural Gradient}
During the training process of DNNs, the purpose is to minimize a loss function $h(\omega)$. Throughout this paper, we use $\mathbb{E}[\cdot]$ to represent the mean of the samples $(x,y)$ and the cross-entropy loss function is computed as
\[ h(\omega)=\mathbb{E}[-\log p(y|x, \omega)],\]
where $\omega$ is a vector of parameters, $x$ is the input, $y$ is the label, and $p(y|x, \omega)$ represents the density function of the neural network's  predictive distribution $P_{y|x}(\omega)$.

Natural gradient was first proposed by \citet{nat1998}. It gives the steepest descent direction in the distribution space rather than the space of parameters. In distribution space, the distance between two distributions $P(\omega)$ and $P(\omega+\Delta\omega)$ is measured by the K-L divergence: ${\rm{D_{KL}}}(P(\omega)\|P(\omega+\Delta\omega))\approx\frac{1}{2}\omega^\top F\omega$, where $F$ is the FIM, and is defined as
\begin{equation}\label{fim}
 F=\mathbb{E}[\nabla_{\omega}\log p(y|x,\omega)\nabla_{\omega}\log p(y|x,\omega)^\top].
\end{equation}

The natural gradient is usually defined as $F^{-1}\nabla_\omega h$, and it provides the update direction for natural gradient descent. So, the parameters are updated by
\begin{equation}\label{nat}
\omega\leftarrow\omega-\eta F^{-1}\nabla_{\omega}h,
\end{equation}
where $\eta$ is the learning rate.

\subsection{KFAC}

For DNNs which have millions or even billions of parameters, it is impractical to compute the exact FIM and its inverse matrix. KFAC provides an useful approximation.  Consider a DNN with $L$ layers and denote the inputs $a_{l-1}$ which are the activations of the previous layer, outputs $s_l$, and weight $W_l$ for the $l$-th layer. Then, we have $s_l=W_la_{l-1}$. For simplicity, we will use the following notation:
\[ \mathcal{D}t:=\nabla_t\log p(y|x,\omega), u_l:=\mathcal{D}s_l,\]
where $t$ is an arbitrary parameter. Therefore, the gradient of weight is $\mathcal{D}W_l=a_{l-1}u_l$, and the Eq. (\ref{fim}) can be written as $F=\mathbb{E}[\mathcal{D}\omega\mathcal{D}\omega^\top]$.

Firstly, KFAC approximates the FIM $F$ as a block diagonal matrix
\begin{equation}\label{kfac1}
  F\approx diag(F_1, F_2, \cdots, F_L)=diag(\mathbb{E}[\mathcal{D}\omega_1\mathcal{D}\omega_1^\top], \mathbb{E}[\mathcal{D}\omega_2\mathcal{D}\omega_2^\top], \cdots, \mathbb{E}[\mathcal{D}\omega_L\mathcal{D}\omega_L^\top]),
\end{equation}
where $\omega_l={\rm{vec}}(W_l)$ for any $l\in\{1, 2, \cdots, L\}$.

Then, each block matrix of the FIM can be written as
\begin{equation}\label{kfac2}
F_l=\mathbb{E}[\mathcal{D}\omega_l\mathcal{D}\omega_l^\top]=\mathbb{E}[(a_{l-1}\otimes u_l)(a_{l-1}\otimes u_l)^\top]\approx\mathbb{E}[a_{l-1}^\top a_{l-1}]\otimes\mathbb{E}[u_l^\top u_l]=A_{l-1}\otimes U_{l},
\end{equation}
where $\otimes$ represents the Kronecker product, $A_{l-1}=\mathbb{E}[a_{l-1}a_{l-1}^\top]$ and $U_l=\mathbb{E}[u_lu_l^\top]$. Due to the properties of Kronecker product $(A_{l-1}\otimes U_l)^{-1}=A_{l-1}^{-1}\otimes U_l^{-1}$ and $(A_{l-1}\otimes U_l){\rm vec}(X)={\rm vec}(U_lXA_{l-1}^\top)$ for any matrix $X$, decomposing $F_l$ into $A_{l-1}$ and $U_l$ not only saves the cost of storing and inverting the exact FIM, but also enables tractable methods to compute the approximate natural gradient
\[
(A_{l-1}\otimes U_l)^{-1}\nabla_{\omega_l}h=(A_{l-1}\otimes U_l)^{-1}{\rm vec}(\nabla_{W_l}\phi)={\rm vec}(U_l^{-1}(\nabla_{W_l}\phi)A_{l-1}^{-1}).
\]

\subsection{EKFAC}
\citet{ekfac2018} proposes an other interpretation of the natural gradient update $F^{-1}\nabla_{\omega}h$. Let $F=Q_F\Lambda_F Q^\top_F$ be the eigendecomposition of the FIM, where $\Lambda$ is a diagonal matrix with eigenvalues and $Q$ is an orthogonal matrix whose columns correspond to eigenvectors. Then, the natural gradient update will be
\begin{equation}\label{nat3}
F^{-1}\nabla_{\omega}h=\underbrace{Q_F\overbrace{\Lambda_F^{-1} \underbrace{Q^\top_F\nabla_{\omega}h}_{\text{(a)}}}^{\text{(b)}}}_{\text{(c)}}.
\end{equation}
The Eq. (\ref{nat3}) can be explained by three steps: (a) multiplying $\nabla_{\omega}h$ by $Q^\top_F$, which projects the gradient vector $\nabla_{\omega}h$ to the eigenbasis $Q_F$; (b) multiplying by the diagonal matrix $\Lambda_F$, which re-scales the coordinates in that eigenbasis by the diagonal inverse matrix $\Lambda_F^{-1}$; (c) multiplying by $Q_F$, which projects the re-scaled coordinates back to the initial basis. The re-scaling factor can be computed by $(\Lambda_F)_{ii}=\mathbb{E}[(Q^\top_F\nabla_{\omega}h)_i^2]$, whose entries are the second moment of the vector $Q^\top_F\nabla_{\omega}h$ (the gradient vector in the eigenbasis). Under this interpretation, for a diagonal approximation of the FIM, the
re-scaling factor is $(\Lambda_F)_{ii}=\mathbb{E}[(\nabla_{\omega}h)_i^2]$ and the eigenbasis can be chosen as the identity matrix $I$. Although the
re-scaling factor is efficient, obtaining an exact eigenbasis is difficult, the eigenbasis $I$ is too crude which leads to great approximation error.

KFAC decomposes the FIM $F$ into two Kronecker factors $A_{l-1}$ and $U_l$. Because $A_{l-1}$ and $U_l$ are real symmetric positive semi-define matrices, they can be expressed as $A_{l-1}=Q_{A_{l-1}}\Lambda_{A_{l-1}} Q^\top_{A_{l-1}}$ and $U_l=Q_{U_l}\Lambda_{U_l} Q^\top_{U_l}$ by eigendecomposition. By the property of Kronecker product, Eq. (\ref{kfac2}) can be written as
\begin{equation}\label{ekfac1}
\begin{aligned}
  F_l&\approx A_{l-1}\otimes U_l=(Q_{A_{l-1}}\Lambda_{A_{l-1}} Q^\top_{A_{l-1}})\otimes(Q_{U_l}\Lambda_{U_l} Q^\top_{U_l})\\
  &=(Q_{A_{l-1}}\otimes Q_{U_l})(\Lambda_{A_{l-1}}\otimes \Lambda_{U_l})(Q_{A_{l-1}}\otimes Q_{U_l})^\top.
\end{aligned}
\end{equation}
According to this interpretation, $Q_{A_{l-1}}\otimes Q_{U_l}$ gives the eigenbasis of the Kronecker product $A_{l-1}\otimes U_l$. Compared with diagonal approximations, KFAC provides a more exact eigenbasis approximation of the full FIM eigenbasis. However, the re-scaling factor is not accurate under the KFAC eigenbasis, that is $(\Lambda_{A_{l-1}}\otimes \Lambda_{U_l})_{ii}\neq \mathbb{E}[((Q_{A_{l-1}}\otimes Q_{U_l})^\top\nabla_{\omega_l}h)_i^2]$. EKFAC corrects this inexact re-scaling factor by defining
\[
(\Lambda_l^\ast)_{ii}=\mathbb{E}[((Q_{A_{l-1}}\otimes Q_{U_l})^\top\nabla_{\omega_l}h)_i^2].
\]
Then, $F_l$ can be approximated as
\begin{equation}\label{ekfac2}
  F_l\approx(Q_{A_{l-1}}\otimes Q_{U_l})\Lambda_l^\ast(Q_{A_{l-1}}\otimes Q_{U_l})^\top.
\end{equation}

\subsection{TKFAC}

Recently, \citet{tkfac2020} proposed a new approximation of $F_l$, which approximates $F_l$ as a Kronecker product scaled by a coefficient $\sigma_l$, i.e.,
\begin{equation}\label{tkfac}
  F_l\approx\sigma_l\Phi_l\otimes \Psi_l,
\end{equation}
where $0<\sigma_l<\infty$ is an unknown parameter, $\Phi_l$ and $\Psi_l$ are two unknown matrices with known traces. Denote $\Lambda_{l-1}=a_{l-1}a_{l-1}^\top$ and $\Gamma_{l}=u_lu_l^\top$. Then, the factors in Eq. (\ref{tkfac}) can be computed by
\begin{equation}\label{tkfac1}
  \sigma_l = \frac{\mathbb{E}[{\rm{tr}}(\Lambda_{l-1}){\rm{tr}}(\Gamma_{l})]}{{\rm tr}(\Phi_l){\rm tr}(\Psi_l)},
  \Phi_l = \frac{{\rm tr}(\Phi_l)\mathbb{E}[{\rm{tr}}(\Gamma_{l})\Lambda_{l-1}]}{\mathbb{E}[{\rm{tr}}
     (\Lambda_{l-1}){\rm{tr}}(\Gamma_{l})]},
  \Psi_l = \frac{{\rm tr}(\Psi_l)\mathbb{E}[{\rm{tr}}(\Lambda_{l-1})\Gamma_{l}]}{\mathbb{E}[{\rm{tr}}(\Lambda_{l-1})
    {\rm{tr}}(\Gamma_{l})]}.
\end{equation}
An important property of TKFAC is to keep the traces equal, i.e., ${\rm{tr}}(F_l)={\rm{tr}}(\sigma_l\Phi_l\otimes\Psi_l)=\sigma_l{\rm{tr}}(\Phi_l){\rm{tr}}(\Psi_l)$. Theoretically, the upper bound of TKFAC's approximation error is smaller than KFAC in general cases. What's more, experimental results show that TKFAC can keep smaller approximation error than KFAC during training. In practice, to reduce computing costs, we can assume that ${\rm{tr}}(\Phi_l)={\rm{tr}}(\Psi_l)=1$. So, Eq. (\ref{tkfac1}) can be simplified as
\begin{equation}\label{tkfac2}
  \sigma_l=\mathbb{E}[{\rm{tr}}(\Lambda_{l-1}){\rm{tr}}(\Gamma_{l})],~
  \Phi_l=\frac{\mathbb{E}[{\rm{tr}}(\Gamma_{l})\Lambda_{l-1}]}{\mathbb{E}[{\rm{tr}}(\Lambda_{l-1}){\rm{tr}}(\Gamma_{l})]},~
  \Psi_l=\frac{\mathbb{E}[{\rm{tr}}(\Lambda_{l-1})\Gamma_{l}]}{\mathbb{E}[{\rm{tr}}(\Lambda_{l-1}){\rm{tr}}(\Gamma_{l})]}.
\end{equation}

\section{Methods} \label{sec-4}
\setcounter{equation}{0}
\subsection{TEKFAC}
EKFAC corrects the inexact re-scaling factor in KFAC based on the model that $F_l$ is approximated the Kronecker product of two smaller matrices. If we think of TKFAC in terms of the interpretation adopted by EKFAC, the re-scaling factor in TKFAC is also inexact. So, in this section, we combine the ideas of these two methods and propose a new method called TEKFAC, which can keep track of the diagonal variance in TKFAC eigenbasis.

TKFAC approximates $F_l$ as a Kronecker product of two factors $\Phi_l,\Psi_l$ and scaled by the coefficient $\sigma_l$. It is easy to know that $\Phi_l$ and $\Psi_l$ are symmetric positive semi-define matrices, according to eigendecomposition, we can obtain
\begin{equation}\label{etkfac1}
  \begin{aligned}
  F_l&\approx\sigma_l\Phi_l\otimes \Psi_l=\sigma_l(Q_{\Phi_l}\Lambda_{\Phi_l} Q^\top_{\Phi_{l}})\otimes(Q_{\Psi_l}\Lambda_{\Psi_l} Q^\top_{\Psi_l})\\
  &=\sigma_l(Q_{\Phi_l}\otimes Q_{\Psi_l})(\Lambda_{\Phi_l}\otimes \Lambda_{\Psi_l})(Q_{\Phi_l}\otimes Q_{\Psi_l})^\top ,
\end{aligned}
\end{equation}
where $\Lambda_{\Phi_l}, \Lambda_{\Psi_l}$ are two diagonal matrices with eigenvalues of $\Phi_l,\Psi_l$ and $Q_{\Phi_l},\Lambda_{\Phi_l}$ are two orthogonal matrices whose columns are eigenvectors of $\Phi_l,\Psi_l$, respectively.
As the interpretation in subsection 3.2, $Q_{\Phi_l}\otimes Q_{\Psi_l}$ gives the TKFAC eigenbasis, and the re-scaling factor can be selected as  $\sigma_l(\Lambda_{\Phi_l}\otimes \Lambda_{\Psi_l})$. However, this re-scaling factor is also not guaranteed to match the second moment of the gradient vector in TKFAC eigenbasis, that is $(\Lambda_{\Phi_l}\otimes \Lambda_{\Psi_l})_{ii}\neq\mathbb{E}[((Q_{\Phi_l}\otimes Q_{\Psi_l})^\top\nabla_{\omega_l}h)_i^2]$. Therefore, combined with the idea of EKFAC, we redefine the re-scaling factor by
\begin{equation}\label{etkfac2}
(\Theta_l)_{ii}=\mathbb{E}[((Q_{\Phi_l}\otimes Q_{\Psi_l})^\top\nabla_{\omega_l}h)_i^2],
\end{equation}
where $\Theta_l$ is a diagonal matrix. Eq. (\ref{etkfac2}) defines a more accurate re-scaling factor. Then, we can obtain the new approxiamtion defined as follows
\begin{equation}\label{etkfac3}
  F_l\approx(Q_{\Phi_l}\otimes Q_{\Psi_l})\Theta_l(Q_{\Phi_l}\otimes Q_{\Psi_l})^\top.
\end{equation}

Similar to the analysis process of EKFAC, we can proof that $\Theta_l$ is the optimal diagonal scaling factor under the TKFAC eigenbasis. That is $\Theta_l$ is the optimal solution to the following problem.
\begin{eqnarray*}
  \min\limits_{\Lambda_l} && \|F_l-(Q_{\Phi_l}\otimes Q_{\Psi_l})\Lambda_l(Q_{\Phi_l}\otimes Q_{\Psi_l})^\top\|_F \\
  s.t. && \Lambda_l\ {\rm  is\ a\ diagonal\ matrix}
\end{eqnarray*}
According to this conclusion, we can easily prove the following theorem. For simplicity, we omit the subscript in the following theorem.
\begin{theorem}
Let $F_{\TKFAC}$ and $F_{\TEKFAC}$ are the approximate matrices of the FIM $F$, i.e.,
\[
  F_{\TKFAC} = (Q_{\Phi}\otimes Q_{\Psi})(\sigma\Lambda_{\Phi}\otimes \Lambda_{\Psi})(Q_{\Phi}\otimes Q_{\Psi})^\top,
\]
\[
  F_{\TEKFAC} = (Q_{\Phi}\otimes Q_{\Psi})\Theta(Q_{\Phi}\otimes Q_{\Psi})^\top,
\]
then, we have $\|F-F_{\TEKFAC}\|_F\leq\|F-F_{\TKFAC}\|_F$.
\end{theorem}
\begin{proof}
Because
\[
\Theta=\arg\min_{\Lambda}\|F-(Q_{\Phi}\otimes Q_{\Psi})\Lambda(Q_{\Phi}\otimes Q_{\Psi})^\top\|_F
\]
and for the diagonal matrices $\Lambda_{\Phi}\otimes \Lambda_{\Psi}$, $\Theta_l$
\[
(\Lambda_{\Phi}\otimes \Lambda_{\Psi})_{ii}\neq\mathbb{E}[((Q_{\Phi}\otimes Q_{\Psi})^\top\nabla_{\omega}h)_i^2]=\Theta_{ii}.
\]
Therefore, we have
\[
\|F-(Q_{\Phi}\otimes Q_{\Psi})\Theta(Q_{\Phi}\otimes Q_{\Psi})^\top\|_F
\leq
\|F-(Q_{\Phi}\otimes Q_{\Psi})(\sigma\Lambda_{\Phi}\otimes \Lambda_{\Psi})(Q_{\Phi}\otimes Q_{\Psi})^\top\|_F
\]
that is
\[
\|F-F_{\TEKFAC}\|_F\leq\|F-F_{\TKFAC}\|_F.
\]
The proof is complete.
\end{proof}
So, TEKFAC provides a more accurate approximation for the FIM than TKFAC in theory. To use the second-order optimization methods effectively in practice, a suitable damping technique is also necessary. Crucially, powerful second-order optimizers like KFAC and EKFAC usually require more complicated damping techniques, otherwise, they will tend to fail completely. KFAC introduces an effective damping technique by adding $\sqrt{\lambda} I$ to the Kronecker factors $A_{l-1}$ and $U_l$. In EKFAC, since the re-scaling factor has been revised and redefined, it is no longer useful to add damping to the Kronecker factors, and the damping should be added to the re-scaling factor.
TKFAC adopts the same damping technique as KFAC for FNNs and proposes a new automatic tuning damping for CNNs. For TEKFAC, we also use the same damping technique as EKFAC for FNNs and the new damping technique adopted in TKFAC for CNNs, i.e.,

\begin{equation}\label{dam1}
F_l\approx(Q_{\Phi_l}\otimes Q_{\Psi_l})(\Theta_l+\lambda I)(Q_{\Phi_l}\otimes Q_{\Psi_l})^\top
\end{equation}
where $\lambda$ is a reasonably large positive scalar for FNNs and

\begin{equation}\label{dam2}
\lambda=\frac{\max\{{\rm tr}(\Theta_l),\vartheta\}}{{\rm dim}(\Theta_l)}
\end{equation}
for CNNs. In Eq. (\ref{dam2}), $\vartheta$ is a reasonably large positive scalar and dim denotes the number of the rows (or columns) of $\Theta_l$. What's more, in order to keep pace with convolution layers, we expanded the FIM of the fully connected layer in CNNs by a factor of $\beta$ as described in \citet{tkfac2020}, where $\beta=\max_{l\in\{\rm convolutional \; layers\}}\left\{\max\{{\rm tr}(\Theta_l),\vartheta\}/{\rm dim}(\Theta_l)\right\}$.
This damping technique for CNNs was first used in \citet{tkfac2020}. The purpose is to dynamically adjust the damping based on the FIM's trace during training, so the damping can be adapted to the FIM's elements to avoid the problem that the damping is large enough to transform the second-order optimizer into the first-order one during training.

\begin{algorithm}[htb]
\caption{TEKFAC algorithm}\label{alg1}

\begin{algorithmic}
\REQUIRE $\eta$ : learning rate
\REQUIRE $\lambda$ : damping parameter
\REQUIRE $\beta_1$ : exponential moving average parameter of the re-scaling factor $\Theta_l$
\REQUIRE $\beta_2$ : exponential moving average parameter of factors $\Phi_l$ and $\Psi_l$
\REQUIRE $T_{\rm{FIM}}, T_{\rm{EIG}}, T_{\rm{RE}}$ : FIM, eigendecomposition and re-scaling update intervals
\STATE   $k\leftarrow0$
\STATE   Initialize $\{\delta_l\}_{l=1}^L, \{\Phi_l\}_{l=1}^L$, $\{\Psi_l\}_{l=1}^L$ and $\{\Theta_l\}_{l=1}^L$

\WHILE{convergence is not reached}

\IF{$k\equiv0$ (mod $T_{\rm{FIM}}$)}
\STATE Update the factors $\{\delta_l\}_{l=1}^L, \{\Phi_l\}_{l=1}^L$ and $\{\Psi_l\}_{l=1}^L$ using Eq. (\ref{tkfac2})
\ENDIF
\IF{$k\equiv0$ (mod $T_{\rm{EIG}}$)}
\STATE Compute the eigenbasis $Q_{\Phi_l}$ and $Q_{\Psi_l}$ using Eq. (\ref{etkfac1}), (\ref{etkfac7}) and (\ref{etkfac8})
\ENDIF
\IF{$k\equiv0$ (mod $T_{\rm{RE}}$)}
\STATE Update the re-scaling factor $\{\Theta_l\}_{l=1}^L$ using Eq. (\ref{etkfac2}), (\ref{dam1}) and (\ref{etkfac6})
\ENDIF

\STATE $\nabla^{(k)}_l\leftarrow (Q_{\Phi_l}^\top\otimes Q_{\Psi_l}^\top)^{(k)} \nabla_{\omega_l} h^{(k)}$
\STATE $\nabla^{(k)}_l\leftarrow \nabla^{(k)}_l/({\rm vec}(\Theta_l^{(k)}+\lambda I))$ (element-wise scaling)
\STATE $\nabla^{(k)}_l\leftarrow (Q_{\Phi_l}\otimes Q_{\Psi_l})^{(k)}\nabla^{(k)}_l$
\STATE $\omega^{(k)}_l\leftarrow\omega^{(k-1)}_l-\eta\nabla^{(k)}_l$
\STATE $k\leftarrow k+1$
\ENDWHILE

\end{algorithmic}
\end{algorithm}

For each layer, EKFAC estimates the Kronecker factors $A_{l-1}, U_l$ and the re-scaling factor $(\Lambda_l^\ast)_{ii}$ using exponential moving average. Similarly, we can obtain the exponential moving average updates for TEKFAC in Eq. (\ref{etkfac3}).
\begin{equation}\label{etkfac6}
  (\Theta_l)_{ii}^{(k+1)}\leftarrow\beta_1(\Theta_l)_{ii}^{(k+1)}+(1-\beta_1)(\Theta_l)_{ii}^{(k)},
\end{equation}
\begin{equation}\label{etkfac7}
  \Phi_l^{(k+1)}\leftarrow \beta_2\Phi_l^{(k+1)}+(1-\beta_2)\Phi_l^{(k)},
\end{equation}
\begin{equation}\label{etkfac8}
  \Psi_l^{(k+1)}\leftarrow \beta_2\Psi_l^{(k+1)}+(1-\beta_2)\Psi_l^{(k)},
\end{equation}
where $\beta_1$ and $\beta_2$ are two exponential moving average parameters of the re-scaling factor and Kronecker factors. Finally, TEKFAC updates the parameters by
\begin{equation}\label{etkfac8}
 \omega_l^{(k+1)} \leftarrow \omega_l^{(k)} -\eta (Q_{\Phi_l}\otimes Q_{\Psi_l})^{(k+1)} [(\Theta_l+\lambda I)^{(k+1)}]^{-1}
 (Q_{\Phi_l}^\top\otimes Q_{\Psi_l}^\top)^{(k+1)} \nabla_{\omega_l} h^{(k+1)}.
\end{equation}

Drawing inspiration of EKFAC and TKFAC, we present TEKFAC. Using TEKFAC for training DNNs mainly involves: a) computing the TEKFAC eigenbasis by eigendecomposition; b) estimating the re-scaling factor $\Theta_l$ as defined in Eq. (\ref{etkfac2}); c)  computing the gradient and updating model's parameters. The full algorithm of TEKFAC is given in Algorithm \ref{alg1}, in which the Kronecker product can be computed efficiently by the following identity: $(A \otimes U)\rm{vec}(X)={vec}(U^\top XA)$.

\subsection{Discussion of different methods}

Because the scale of the curvature matrix for DNNs is too large, it is impractical to compute the exact curvature matrix and its inverse matrix for DNNs. In order to effectively use natural gradient descent in training DNNs, KFAC was firstly proposed in \citep{kfac2015}, then EKFAC \citep{ekfac2018} and TKFAC \citep{tkfac2020} were presented gradually. In the last subsection, we propose TEKFAC. In this subsection, we will discuss the relationships and differences of these methods.

\begin{figure}[htb]
  \centering
   \includegraphics[width=0.95\linewidth]{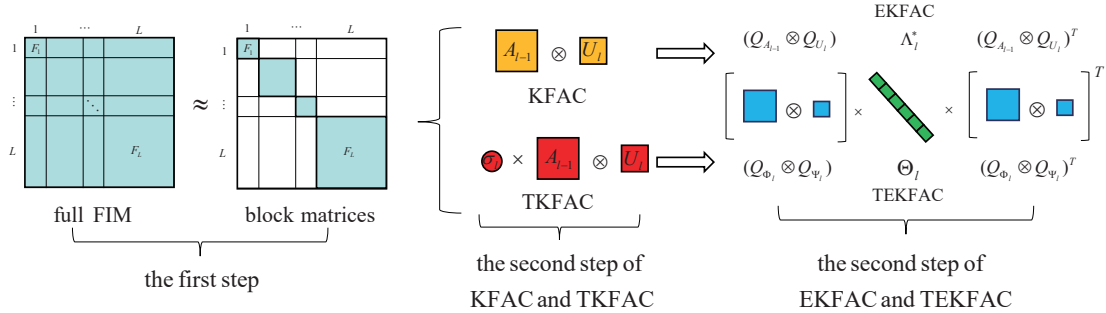}
  \caption{Illustration of the approximation process of KFAC, EKFAC, TKFAC and TEKFAC.}\label{ill}
\end{figure}

The approximation process of these methods can be divided into two steps. In the first step, they all decompose the FIM into block matrices according to layers of DNNs. By assuming that parameters of different layers are independent, the inverse of the full FIM
is simplified as the inverse of these small block matrices. This step doesn't make any difference for all these methods. In the second step, KFAC approximates different block matrices as the Kronecker product of two much smaller matrices, EKFAC reinterprets the KFAC by eigenvalue decomposition and corrects the inaccurate re-scaling factor under the KFAC eigenbasis, TKFAC approximates different block matrices as a Kronecker product scaled by a coefficient, TEKFAC corrects the inaccurate re-scaling factor under the TKFAC eigenbasis based on the ideas of EKFAC. The two approximate processes of these methods are illustrated in Figure \ref{ill}. We also summarize the different approximate models and re-scaling factors of these methods in Table \ref{tabdif}.

\begin{table}[H]
\setlength{\abovecaptionskip}{0.cm}
\setlength{\belowcaptionskip}{0.3cm}
\centering
\caption{Summary of some optimizers}
\label{tabdif}
\begin{tabular}{c|c|c}
\toprule
\textbf{optimizer} & \textbf{$F_l$}       & \textbf{re-scaling factor}               \\ \midrule
KFAC\citep{kfac2015}              & $A_{l-1}\otimes U_l$ & $\Lambda_{A_{l-1}}\otimes \Lambda_{U_l}$ \\
EKFAC\citep{ekfac2018}  & $A_{l-1}\otimes U_l$           & diag$(\mathbb{E}[((Q_{A_{l-1}}\otimes Q_{U_l})^\top\nabla_{\omega_l}h)^2])$   \\
TKFAC\citep{tkfac2020}  & $\sigma_l\Phi_l\otimes \Psi_l$ & $\sigma_l(\Lambda_{\Phi_l}\otimes \Lambda_{\Psi_l})$                        \\
TEKFAC & $\sigma_l\Phi_l\otimes \Psi_l$ & diag$(\mathbb{E}[((Q_{\Phi_l}\otimes Q_{\Psi_l})^\top\nabla_{\omega_l}h)^2])$ \\ \bottomrule
\end{tabular}
\end{table}

In TKFAC, an important property is to keep the traces equal before and after the approximation. For TEKFAC, this property can be still kept because
\begin{align*}
{\rm tr}(F_l^{{\rm (TEKFAC)}})&={\rm tr}((Q_{\Phi_l}\otimes Q_{\Psi_l})\Theta_l(Q_{\Phi_l}\otimes Q_{\Psi_l})^\top)={\rm tr}(\Theta_l)\\
&= \sum_i(\Theta_l)_{ii}=\sum_i\mathbb{E}[((Q_{\Phi_l}\otimes Q_{\Psi_l})^\top\nabla_{\omega_l}h)_i^2]\\
&={\rm tr}(\mathbb{E}[(Q_A \otimes Q_U)^\top\nabla_{\omega}h(\nabla_{\omega}h)^\top (Q_A \otimes Q_U)])\\
&={\rm tr}(\mathbb{E}[\nabla_{\omega}h(\nabla_{\omega}h)^\top])={\rm tr}(F_l),
\end{align*}
where $F_l^{{\rm (TEKFAC)}}$ represents the approximation defined by Eq. (\ref{etkfac3}) and $F_l$ is the exact FIM. Similar to this conclusion, EKFAC can also keep the traces equal. However, we should note that EKFAC is based on the KFAC (Eq. (\ref{kfac2})) and correcting the re-scaling factor, then the traces can be kept equal. TKFAC proposes a different approximation (Eq. (\ref{tkfac})) and uses a trace operator to get the calculation formula under the condition that the trace is equal. The motivations for EKFAC and TKFAC are different. Finally, the relationships among these methods are summarized in Figure 4.
\begin{wrapfigure}{r}{0.5\textwidth}
\vspace{-20pt}
\begin{center}
\includegraphics[width=0.5\textwidth]{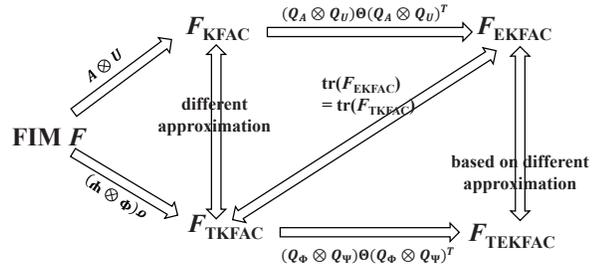}
\end{center}
\vspace{-10pt}
\caption{Illustration of the relationships of KFAC, EKFAC, TKFAC and TEKFAC.}\label{rel}
\end{wrapfigure}

\section{Experiments}\label{sec-5}

To show the effectiveness of TEKFAC, we empirically demonstrate its performance on several standard benchmark datasets for some deep CNNs. Experimental results are given in the following subsection.

\subsection{Setup}

{\textbf{Datasets and models:}} In this paper, we employ three commonly used image classification datasets: CIFAR-10, CIFAR-100 \citep{data2009} and SVHN \citep{data2011}. These datasets all consist of colored images with $32\times 32$ pixels. More details of these datasets are described in Table \ref{tabdata}. We adopt a standard data augmentation scheme including random crop and horizontal flip for CIFAR-10/100, and we do not use data augmentation for SVHN. We consider the performance of different methods on two widely used deep CNNs: VGG16 \citep{net2014} and ResNet20 \citep{net2016}.

\begin{table}[htb]
\setlength{\abovecaptionskip}{0.0cm}
\setlength{\belowcaptionskip}{0.3cm}
\centering
\caption{Statistics of the datasets used in experiments.}
\label{tabdata}
\begin{tabular}{@{}c|c|c|c@{}}
\toprule
\textbf{Dataset} & \textbf{$\#$classes} & \textbf{$\#$training set} & \textbf{$\#$testing set} \\ \midrule
CIFAR-10         & 10               & 50000             & 10000            \\
CIFAR-100        & 100              & 50000             & 10000            \\
SVHN             & 10               & 73257             & 26032            \\ \bottomrule
\end{tabular}
\end{table}

\noindent{\textbf{Baselines and hyper-parameters selection:}} Our method mainly modify the EKFAC eigenbasis according to the model adopted in TKFAC, so we mainly focus on the performance of TEKFAC compared with EKFAC and TKFAC. Therefore, we choose SGDM, Adam, EKFAC and TKFAC as baselines. We mainly refer to the parameters setting \footnote{https://github.com/pomonam/NoisyNaturalGradient}
\footnote{https://github.com/gd-zhang/Weight-Decay} in recent related articles \citep{nekfac2019,wd2019,tkfac2020}. For all experiments, the hyper-parameters are tuned as follows:
\begin{itemize}
  \setlength{\itemsep}{3pt}
  \setlength{\parsep}{0pt}
  \setlength{\parskip}{0pt}
  \item learning rate $\eta$: \{1e-4, 3e-4, 1e-3, 3e-3, 1e-2, 3e-2, 1e-1, 3e-1, 1, 3\}. The initial learning rate is multiplied by 0.1 every 20 epochs for SVHN and every 40 epochs for CIFAR10/100.
  \item damping $\lambda$: \{1e-8, 1e-6, 1e-4, 3e-4, 1e-3, 3e-3, 1e-2, 3e-2, 1e-1, 3e-1\}.
  \item the parameter to restrict trace $\vartheta$: \{1e-4, 1e-3, 1e-2, 1e-1, 1, 10, 100\}.
  \item moving average parameter$\beta_1,\beta_2$ : $\beta_1=\beta_2=0.95$.
  \item momentum: 0.9.
  \item $T_{\rm{FIM}}=T_{\rm{EIG}}=50, T_{\rm{INV}}=200$.
  \item batch size:
  128 for SVHN, CIFAR-10/100.
\end{itemize}

For all methods, we use batch normalization and don't use weight decay. All experiments are run on a single RTX 2080Ti GPU using TensorFlow and repeated three times.

\subsection{Results of experiments}

\begin{figure}[H]
  \centering
  \subfigure[VGG16 on SVHN]{
  \includegraphics[width=0.3\linewidth]{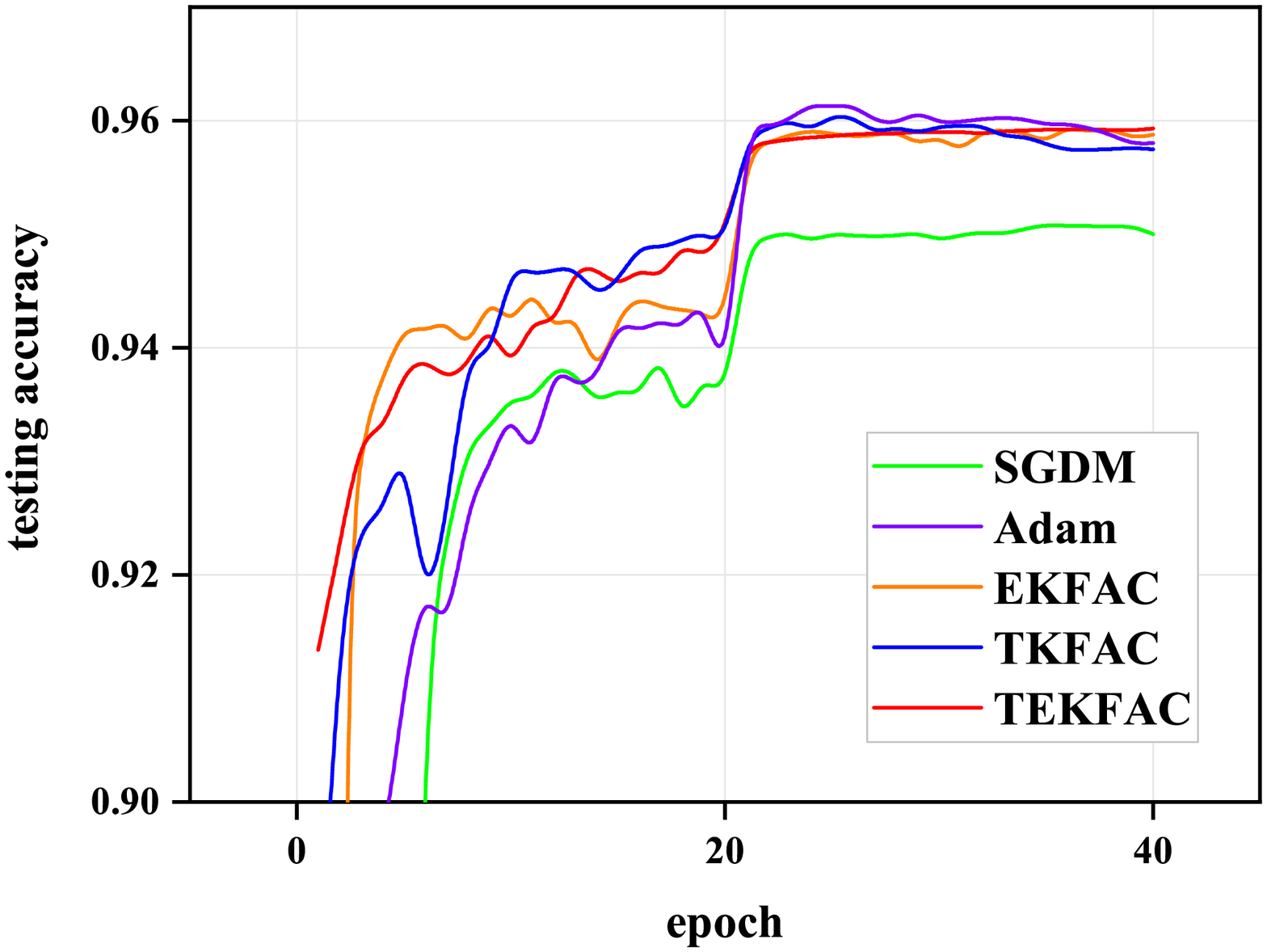}}
  \subfigure[VGG16 on CIFAR-10]{
  \includegraphics[width=0.3\linewidth]{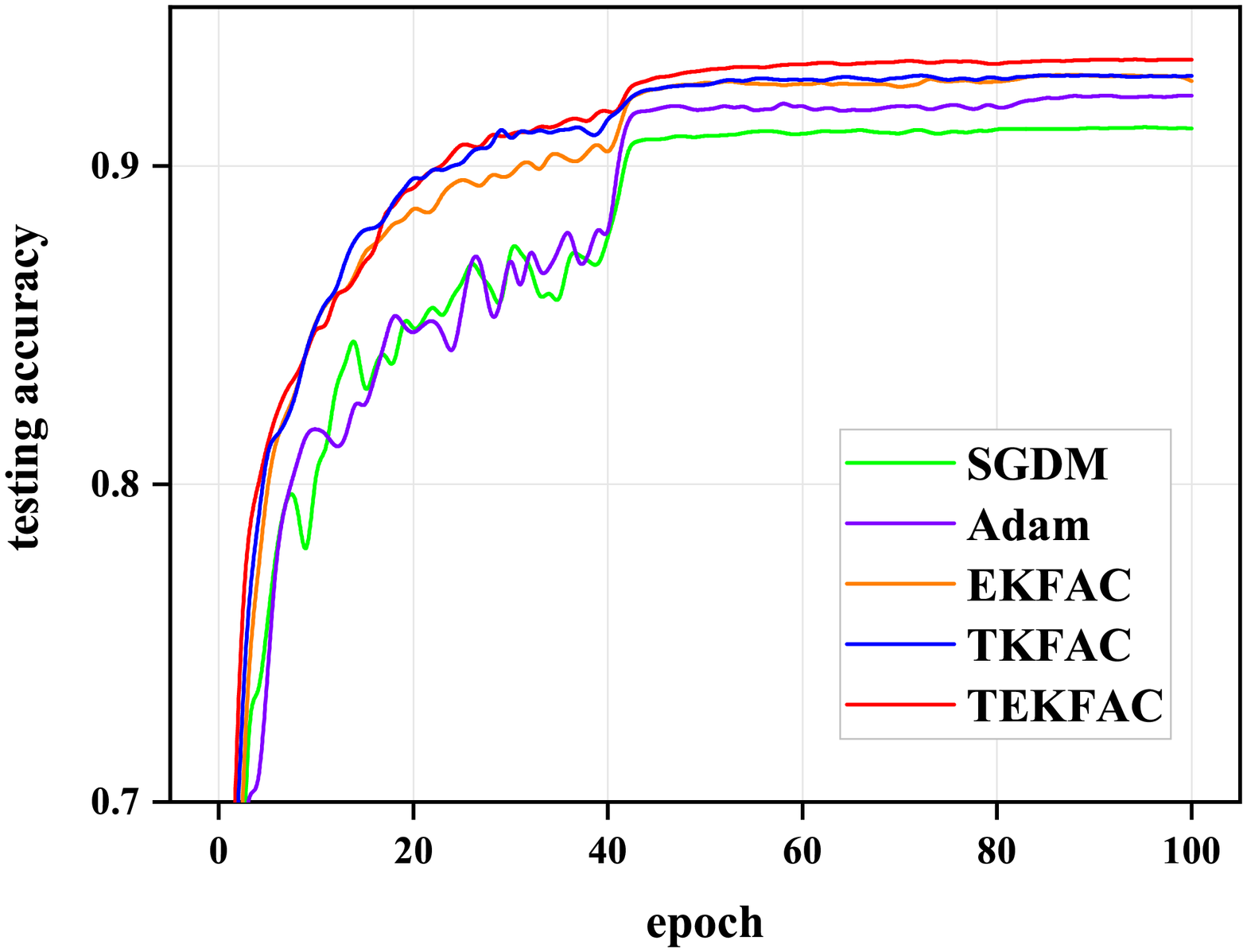}}
  \subfigure[VGG16 on CIFAR-100]{
  \includegraphics[width=0.3\linewidth]{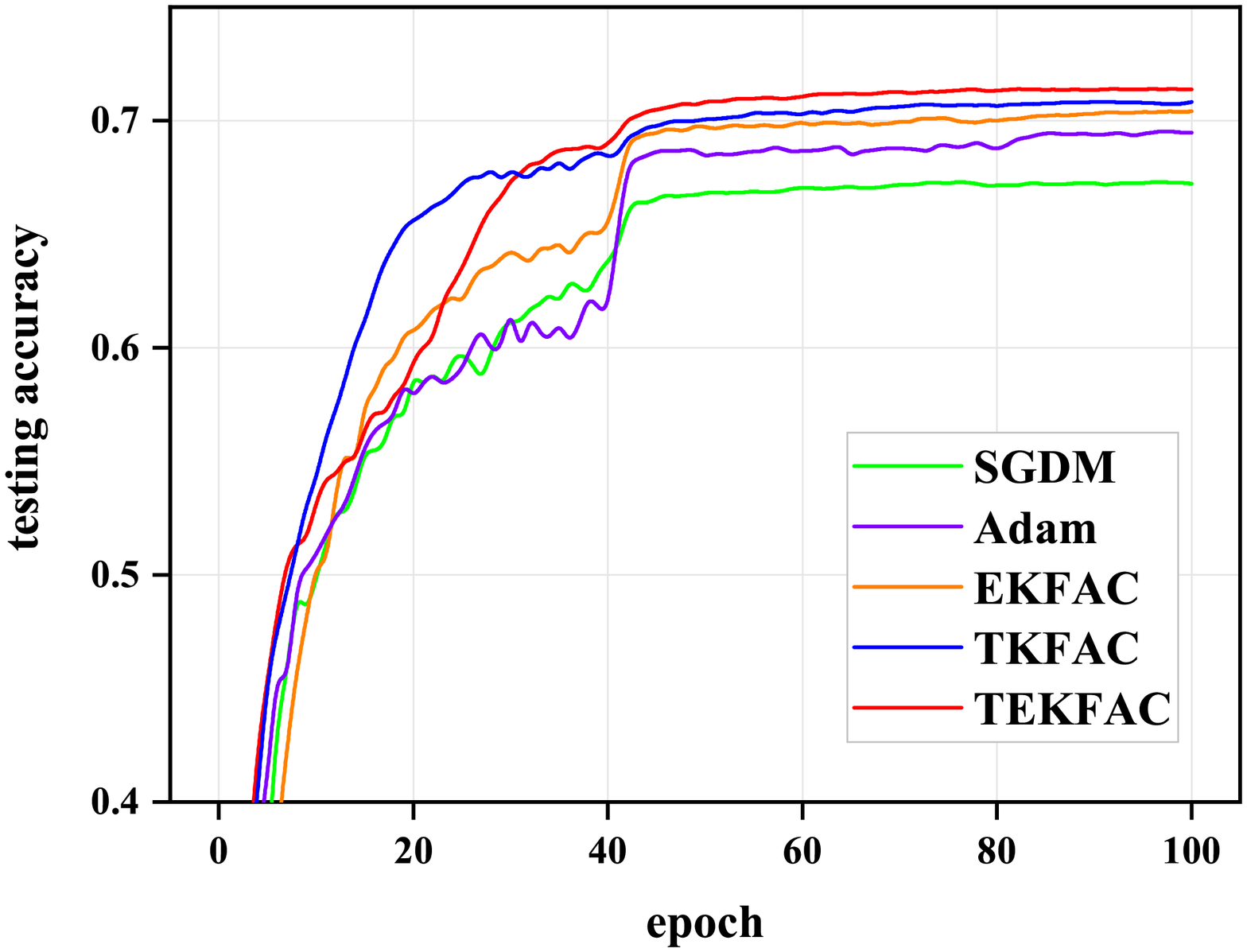}}\\
  \subfigure[ResNet20 on SVHN]{
  \includegraphics[width=0.3\linewidth]{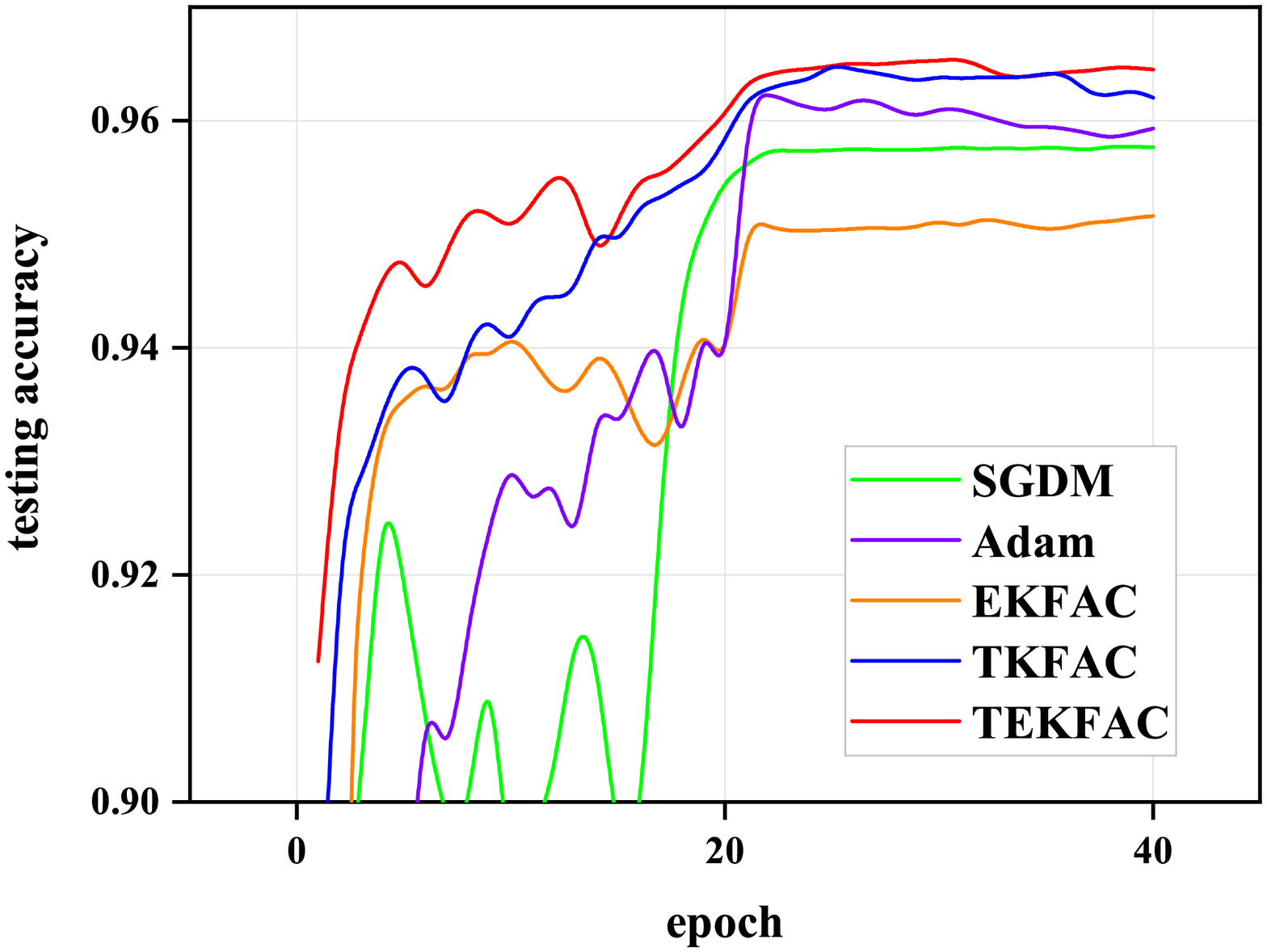}}
  \subfigure[ResNet20 on CIFAR-10]{
  \includegraphics[width=0.3\linewidth]{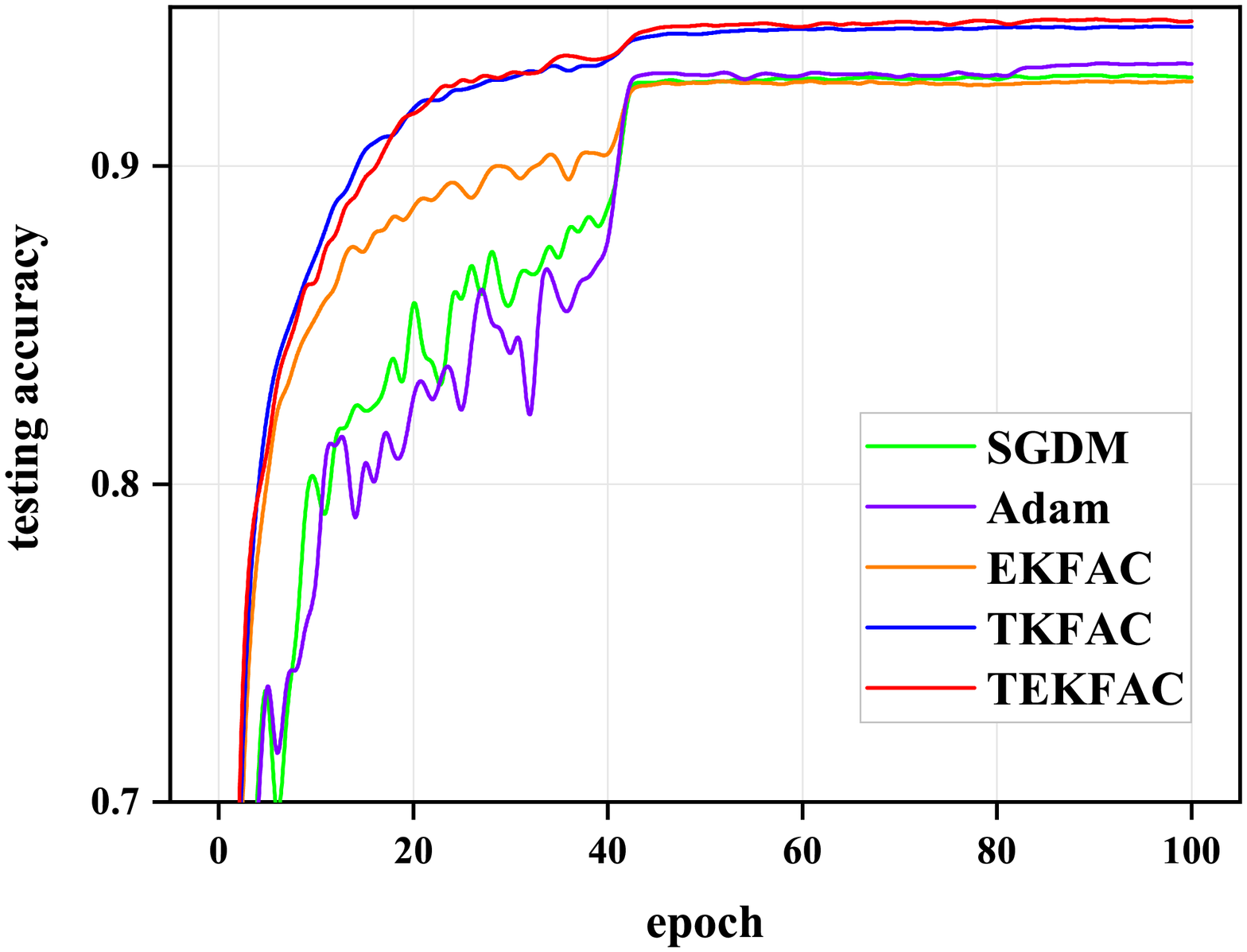}}
  \subfigure[ResNet20 on CIFAR-100]{
  \includegraphics[width=0.3\linewidth]{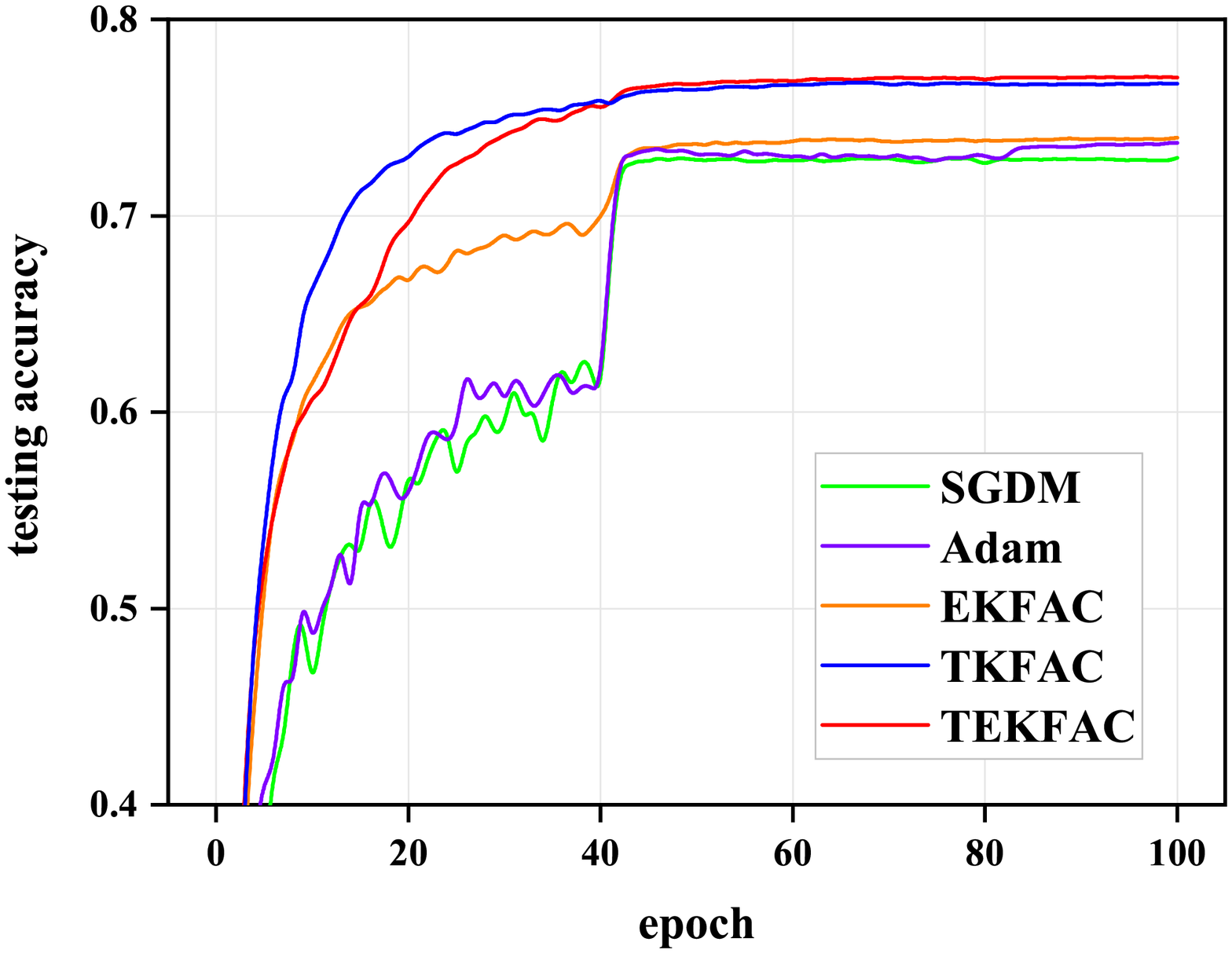}}
  \caption{The curves of testing accuracy with epochs for SGDM, Adam, EKFAC, TKFAC and TEKFAC on SVHN, CIFAR-10 and CIFAR-100. The models we used here are VGG16 and ResNet20. All results are repeated three runs and the curves show the average results. }\label{picacc}
\end{figure}

\noindent{\textbf{Results of CIFAR-10, CIFAR-100 and SVHN: }}We perform extensive experiments on three standard datasets to investigate the effectiveness of TEKFAC. The main results on SVHN and CIFAR10/100 are shown in Figure \ref{picacc} and Table \ref{tabacc}. Figure \ref{picacc} shows the results of SGDM, Adam, EKFAC, TKFAC and TEKFAC on these three datasets in terms of testing accuracy. In Figure \ref{picacc} , we can see that all the second order optimizers (EKFAC, TKFAC and TEKFAC) converge faster than SGDM and Adam. On SVHN and CIFAR-10, TEKFAC achieves same or faster convergence as TKFAC (faster than EKFAC clearly on all datasets) while achieving better accuracy. On CIFAR-100, although TEKFAC converges slower in the first few epochs, it can achieve same convergence as TKFAC after about 30 epochs with better accuracy. The final testing accuracies are summarized in Table \ref{tabacc}.

Table \ref{tabacc} illustrates the testing accuracies of various methods (SGDM, Adam, EKFAC, TKFAC and TEKFAC) with different models (VGG16 and ResNet20) on the SVHN , CIFAR-10 and CIFAR-100 datasets. These experiments are repeated for three times and the results are reported in mean $\pm$ standard deviation. As shown in Table \ref{tabacc}, TEKFAC can achieve higher average accuracy than other baselines in all cases. Compared with EKFAC, TEKFAC greatly improves the testing accuracies of all datasets. For example,  TEKFAC improves 0.96\% and 3.06\% than EKFAC on the CIFAR-100 dataset. Compared with TKFAC, TEKFAC is also able to improve the testing accuracies. For TEKFAC, on the one hand, the idea of EKFAC is combined to correct the inexact re-scaling factor; on the other hand, the new approximation method and the effective damping technique proposed in TKFAC are considered, so a more effective algorithm is obtained. These results also illustrate this point.

\begin{table}[htb]
\small
\setlength{\abovecaptionskip}{0.0cm}
\setlength{\belowcaptionskip}{0.3cm}
\centering
\caption{Results of the SVHN, CIFAR-10 and CIFAR-100 datasets on VGG16 and ResNet20 for SGDM, Adam, EKFAC, TKFAC and TEKFAC. We give the final testing accuracies  (mean $\pm$ standard deviation over three runs)  after 40 epochs for SVHN and 100 epochs for CIFAR. }
\label{tabacc}
\begin{tabular}{@{}cc|ccccc@{}}
\toprule
Dataset   & Model    & SGDM            & Adam            & EKFAC           & TKFAC           & TEKFAC                   \\ \midrule
SVHN      & VGG16    & 94.98$\pm$ 0.09 & 95.80$\pm$ 0.11 & 95.87$\pm$ 0.13 & 95.75$\pm$ 0.21 & \textbf{95.93$\pm$ 0.16} \\
SVHN      & ResNet20 & 95.78$\pm$ 0.15 & 95.93$\pm$ 0.12 & 95.16$\pm$ 0.07 & 96.20$\pm$ 0.39 & \textbf{96.45$\pm$ 0.14} \\
CIFAR-10  & VGG16    & 91.19$\pm$ 0.15 & 92.21$\pm$ 0.14 & 92.67$\pm$ 0.22 & 92.83$\pm$ 0.13 & \textbf{93.35$\pm$ 0.17} \\
CIFAR-10  & ResNet20 & 92.79$\pm$ 0.14 & 93.22$\pm$ 0.18 & 92.66$\pm$ 0.17 & 94.38$\pm$ 0.04 & \textbf{94.56$\pm$ 0.12} \\
CIFAR-100 & VGG16    & 67.29$\pm$ 0.28 & 69.47$\pm$ 0.25 & 70.41$\pm$ 0.26 & 70.82$\pm$ 0.12 & \textbf{71.37$\pm$ 0.18} \\
CIFAR-100 & ResNet20 & 72.94$\pm$ 0.11 & 73.70$\pm$ 0.18 & 73.98$\pm$ 0.21 & 76.73$\pm$ 0.18 & \textbf{77.04$\pm$ 0.15} \\ \bottomrule
\end{tabular}
\end{table}

\noindent{\textbf{Sensitivity to hyper-parameters:}} We also consider the performance of TEKFAC with different hyper-parameters. For TEKFAC, a parameter $\vartheta$ is added to avoid the traces becoming too small during training as TKFAC, so the parameter $\vartheta$ needs to be tuned during traing. Therefore, we mainly consider the effect of the learning rate $\eta$ and the parameter $\vartheta$. We present the results of TEKFAC with different settings on CIFAR-100 with ResNet20.

\begin{table}[htb]
\setlength{\abovecaptionskip}{0.0cm}
\setlength{\belowcaptionskip}{0.3cm}
\centering
\caption{Testing accuracies of different parameter $\vartheta$ on CIFAR-100 with ResNet20 for TEKFAC.}
\label{parameter}
\begin{tabular}{@{}c|cccc@{}}
\toprule
Parameter $\vartheta$ & 0.0001          & 0.001           & 0.01                     & 0.1             \\ \midrule
Testing accuracy                                  & 76.81$\pm$ 0.28 & 76.65$\pm$ 0.12 & \textbf{77.04$\pm$ 0.15} & 76.71$\pm$ 0.17 \\ \midrule
Parameter $\vartheta$ & 1               & 10              & 100                      &                 \\ \midrule
Testing accuracy                                  & 76.03$\pm$ 0.26 & 74.74$\pm$ 0.11 & 73.68$\pm$ 0.09          &                 \\ \bottomrule
\end{tabular}
\end{table}

\begin{figure}[htb]
  \centering
  \subfigure[]{
  \includegraphics[width=0.32\linewidth]{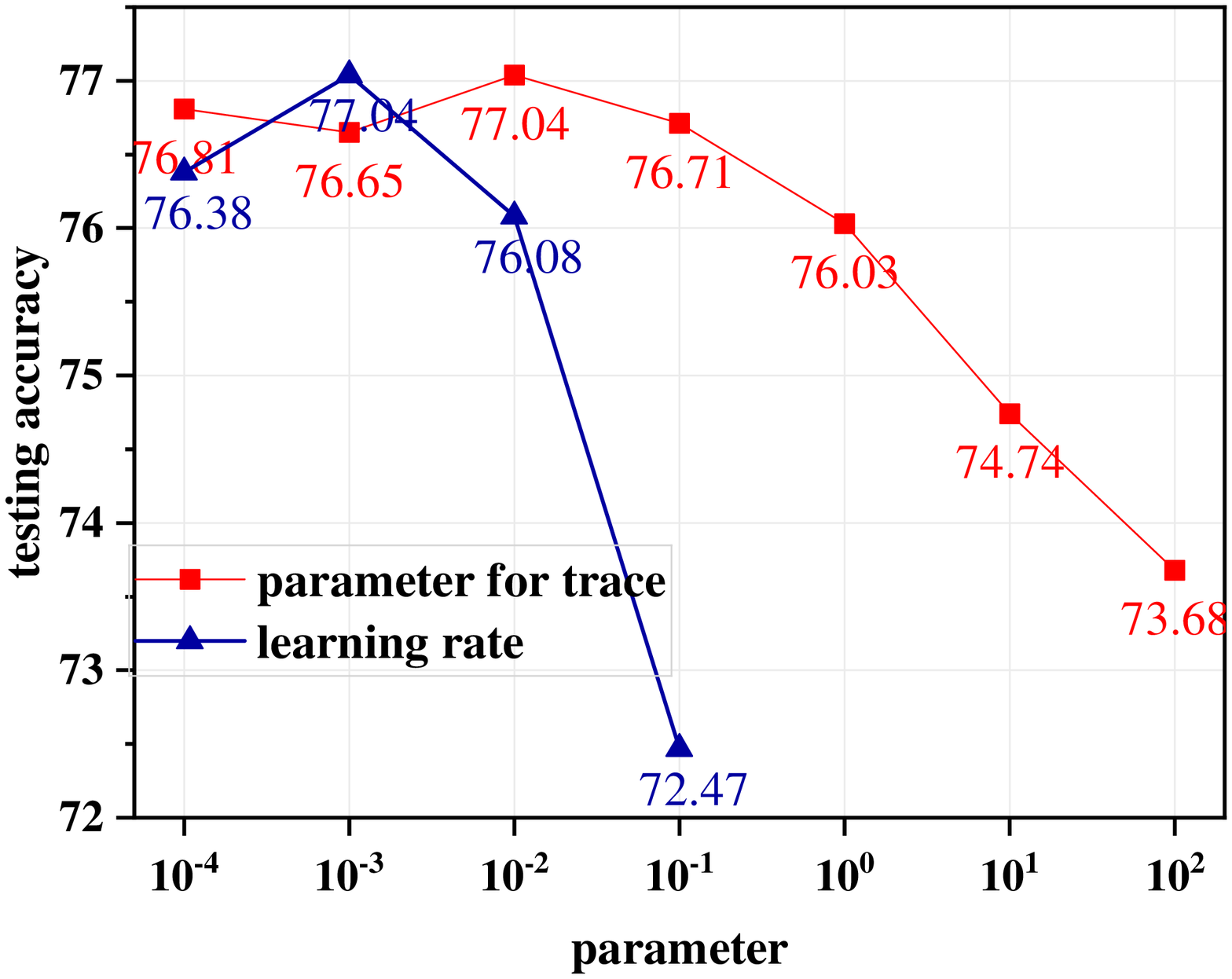}}
  \subfigure[]{
  \includegraphics[width=0.33\linewidth]{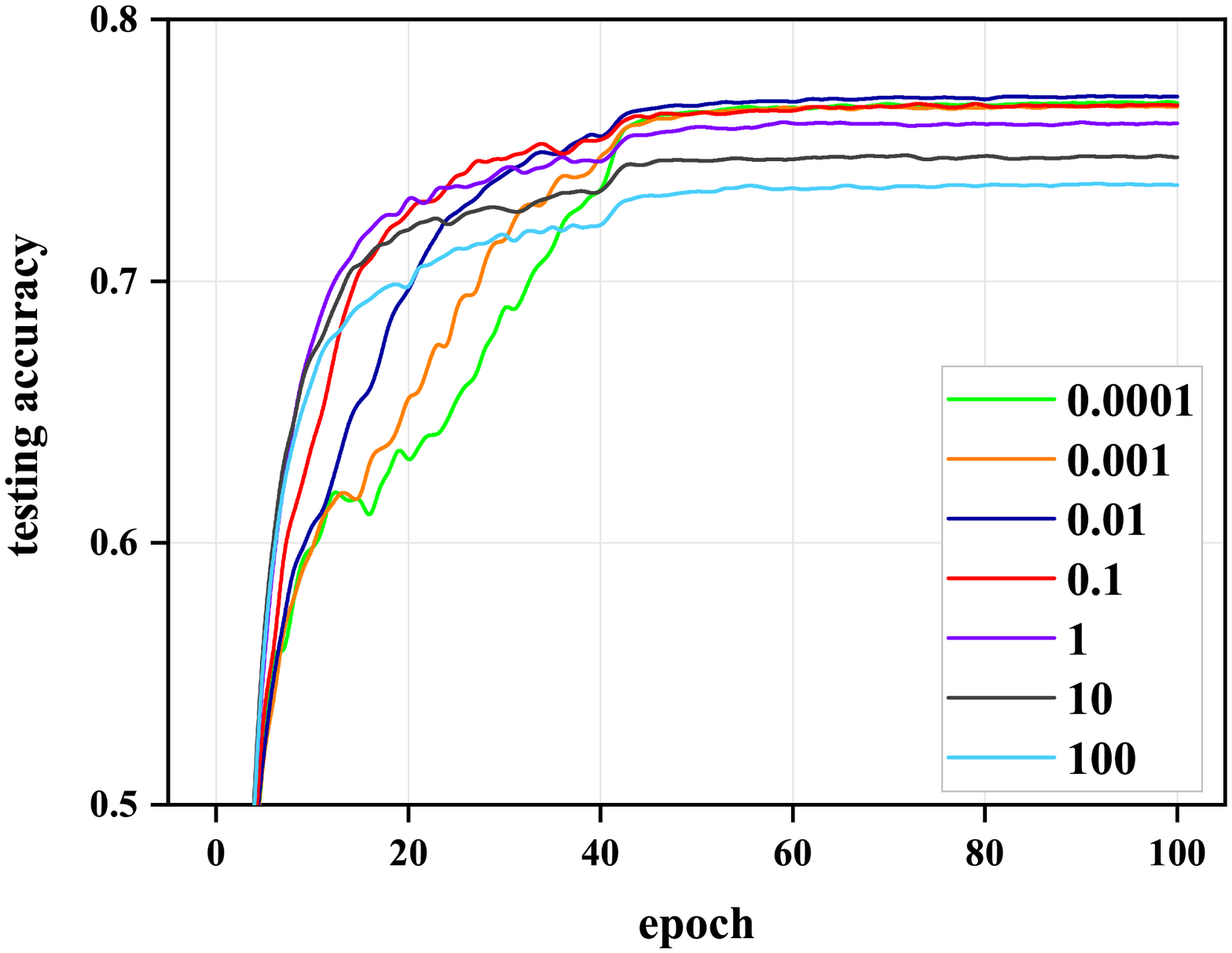}}
  \caption{Results of different parameters. (a) The final testing accuracies of different parameter $\vartheta$ and learning rate $\eta$ with epochs for TEKFAC; (b) The curves of testing accuracies of different $\vartheta$ with epochs for TEKFAC. The $\vartheta$ is set to 0.0001, 0.001, 0.01, 0.1, 1, 10, 100 and $\eta$ is set to 0.0001, 0.001, 0.01, 0.1.} \label{picpra}
\end{figure}

Table \ref{parameter} shows the testing accuracies with different settings of $\vartheta$, where $\vartheta$ is set to 0.0001, 0.001, 0.01, 0.1, 1, 10 and 100, respectively. The learning rate set to 0.001. Figure \ref{picpra} shows the curves of the testing accuracies with epochs for different $\vartheta$. It is clear that the final testing accuracy is similar when $\vartheta \in \{0.0001, 0.001, 0.01, 0.1\}$. However, the testing accuracy decreases rapidly when $\vartheta \geq 1$. On the other hand, we can see that $\vartheta$ also affects the speed of training from Figure \ref{picpra}. When $\vartheta \in \{0.0001, 0.001, 0.01, 0.1, 1\}$, TEKFAC converges slower but may have higher accuracy if $\vartheta$ is smaller. When $\vartheta \in \{10, 100\}$, TEKFAC converges slowly and has lower accuracy. Therefore, we need to select $\vartheta$ carefully to achieve good performance with the balance of training speed and final accuracy. For example, we choose $\vartheta=0.01$ on ResNet20 in this paper. Of course, 0.01 is not suitable for all networks, and $\vartheta$ should be changed for different DNNs.

Table \ref{parameter} shows the testing accuracies with different settings of the learning rate $\eta$, where $\eta$ is set to 0.0001, 0.001, 0.01, and 0.1, respectively. The parameter $\vartheta$ is set to 0.01. We can see that the learning rate also has a great influence on the results of TEKFAC. For CIFAR-100 on ResNet20, 0.001 is a good selection.

\begin{table}[htb]
\setlength{\abovecaptionskip}{0.0cm}
\setlength{\belowcaptionskip}{0.3cm}
\centering
\caption{Testing accuracies of different learning rate $\eta$ on CIFAR-100 with ResNet20 for TEKFAC.}
\label{rate}
\begin{tabular}{@{}c|cccc@{}}
\toprule
Learning tate $\eta$ & 0.0001          & 0.001                    & 0.01            & 0.1                         \\ \midrule
Testing accuracy     & 76.38$\pm$ 0.15 & \textbf{77.04$\pm$ 0.15} & 76.08$\pm$ 0.14 & 72.47$\pm$ 0.23  \\ \bottomrule
\end{tabular}
\end{table}

\section{Conclusions}\label{sec-6}

Inspired by the idea of EKFAC and the new approximation of natural gradient adopted by TKFAC, we proposed TEKFAC algorithm in this work. It not only corrected the inexact re-scaling factor under the TKFAC eigenbasis but also changed the EKFAC eigenbasis based on the new approximation. The relationships of recent methods have also been discussed. Experimental results showed that our method outperformed SGDM, Adam, EKFAC and TKFAC. Of course, the performance of our method on other DNNs or more complex large-scale training tasks needs to be further studied.

\bibliographystyle{plainnat}
\bibliography{d}

\begin{thebibliography}{30}
\providecommand{\natexlab}[1]{#1}
\providecommand{\url}[1]{\texttt{#1}}
\expandafter\ifx\csname urlstyle\endcsname\relax
  \providecommand{\doi}[1]{doi: #1}\else
  \providecommand{\doi}{doi: \begingroup \urlstyle{rm}\Url}\fi

\bibitem[Amari(1998)]{nat1998}
Shun-Ichi Amari.
\newblock Natural gradient works efficiently in learning.
\newblock \emph{Neural Computation}, 10\penalty0 (2):\penalty0 251--276, 1998.

\bibitem[Ba et~al.(2017)Ba, Grosse, and Martens]{dis2017}
Jimmy Ba, Roger Grosse, and James Martens.
\newblock Distributed second-order optimization using kronecker-factored
  approximations.
\newblock In \emph{International Conference on Learning Representations}, 2017.

\bibitem[Bae et~al.(2018)Bae, Zhang, and Grosse]{nekfac2019}
Juhan Bae, Guodong Zhang, and Roger Grosse.
\newblock Eigenvalue corrected noisy natural gradient.
\newblock In \emph{Workshop of Bayesian Deep Learning, Advances in Neural
  Information Processing Systems}, 2018.

\bibitem[Berahas et~al.(2019)Berahas, Jahani, and Tak{\'a}{\v{c}}]{qn2019}
Albert~S Berahas, Majid Jahani, and Martin Tak{\'a}{\v{c}}.
\newblock Quasi-{Newton} methods for deep learning: Forget the past, just
  sample.
\newblock \emph{arXiv preprint arXiv:1901.09997}, 2019.

\bibitem[Dennis and Mor\'e(1977)]{qn1977}
J.~E. Dennis and Jorge~J. Mor\'e.
\newblock Quasi-{Newton} methods, motivation and theory.
\newblock \emph{SIAM Review}, 19\penalty0 (1):\penalty0 46--89, 1977.

\bibitem[Duchi et~al.(2011)Duchi, Elad, and Yoram]{ada2011}
John Duchi, Hazan Elad, and Singer Yoram.
\newblock Adaptive subgradient methods for online learning and stochastic
  optimization.
\newblock \emph{Journal of Machine Learning Research}, 12\penalty0
  (Jul):\penalty0 2121--2159, 2011.

\bibitem[Gao et~al.(2020)Gao, Liu, Huang, Wang, Wang, Xu, and Yu]{tkfac2020}
Kaixin Gao, Xiaolei Liu, Zhenghai Huang, Min Wang, Zidong Wang, Dachuan Xu, and
  Fan Yu.
\newblock Trace-restricted kronecker factorization to approximate natural
  gradient descent for convolution neural networks.
\newblock \emph{arXiv preprint arXiv:2011.10741}, 2020.

\bibitem[George et~al.(2018)George, Laurent, Bouthillier, Ballas, and
  Vincent]{ekfac2018}
Thomas George, C{\'e}sar Laurent, Xavier Bouthillier, Nicolas Ballas, and
  Pascal Vincent.
\newblock Fast approximate natural gradient descent in a kronecker factored
  eigenbasis.
\newblock In \emph{Advances in Neural Information Processing Systems}, pages
  9550--9560, 2018.

\bibitem[Goldfarb et~al.(2020)Goldfarb, Ren, and Bahamou]{qn2020}
Donald Goldfarb, Yi~Ren, and Achraf Bahamou.
\newblock Practical quasi-{Newton} methods for training deep neural networks.
\newblock \emph{arXiv preprint arXiv: 2006.08877v1}, 2020.

\bibitem[Grosse and Martens(2016)]{kfc2016}
Roger Grosse and James Martens.
\newblock A kronecker-factored approximate fisher matrix for convolution
  layers.
\newblock In \emph{International Conference on Machine Learning}, pages
  573--582, 2016.

\bibitem[He et~al.(2016)He, Zhang, Ren, and Sun]{net2016}
Kaiming He, Xiangyu Zhang, Shaoqing Ren, and Jian Sun.
\newblock Deep residual learning for image recognition.
\newblock In \emph{Proceedings of the IEEE Conference on Computer Vision and
  Pattern Recognition}, pages 770--778, 2016.

\bibitem[Kingma and Ba(2014)]{ada2014}
Diederik~P Kingma and Jimmy Ba.
\newblock Adam: A method for stochastic optimization.
\newblock In \emph{International Conference on Learning Representations}, 2014.

\bibitem[Kiros(2013)]{hf2013}
Ryan Kiros.
\newblock Training neural networks with stochastic {Hessian}-free optimization.
\newblock In \emph{International Conference on Learning Representations}, 2013.

\bibitem[Krizhevsky et~al.(2009)Krizhevsky, Hinton, et~al.]{data2009}
Alex Krizhevsky, Geoffrey Hinton, et~al.
\newblock Learning multiple layers of features from tiny images.
\newblock 2009.

\bibitem[Le et~al.(2011)Le, Ngiam, Coates, Lahiri, Prochnow, and Ng]{qn2011}
Quoc~V Le, Jiquan Ngiam, Adam Coates, Abhik Lahiri, Bobby Prochnow, and
  Andrew~Y Ng.
\newblock On optimization methods for deep learning.
\newblock In \emph{International Conference on Machine Learning}, pages
  265--272, 2011.

\bibitem[Martens(2010)]{hf2010}
James Martens.
\newblock Deep learning via {Hessian}-free optimization.
\newblock In \emph{International Conference on Machine Learning}, pages
  735--742, 2010.

\bibitem[Martens and Grosse(2015)]{kfac2015}
James Martens and Roger Grosse.
\newblock Optimizing neural networks with kronecker-factored approximate
  curvature.
\newblock In \emph{International Conference on Machine Learning}, pages
  2408--2417, 2015.

\bibitem[Martens et~al.(2018)Martens, Ba, and Johnson]{kfacr2018}
James Martens, Jimmy Ba, and Matt Johnson.
\newblock Kronecker-factored curvature approximations for recurrent neural
  networks.
\newblock In \emph{International Conference on Learning Representations}, 2018.

\bibitem[Nesterov(1983)]{nes1983}
Yurii Nesterov.
\newblock A method for solving the convex programming problem with convergence
  rate ${O}(1/k^2)$.
\newblock \emph{Soviet Mathematics Doklady}, 27\penalty0 (2):\penalty0
  372--376, 1983.

\bibitem[Netzer et~al.(2011)Netzer, Wang, Coates, Bissacco, Wu, and
  Y~Ng]{data2011}
Yuval Netzer, Tao Wang, Adam Coates, Alessandro Bissacco, Bo~Wu, and Andrew
  Y~Ng.
\newblock Reading digits in natural images with unsupervised feature learning.
\newblock In \emph{NIPS 2011 Workshop on Deep Learning and Unsupervised Feature
  Learning}, 2011.

\bibitem[Osawa et~al.(2019)Osawa, Tsuji, Ueno, Naruse, Yokota, and
  Matsuoka]{lar2019}
Kazuki Osawa, Yohei Tsuji, Yuichiro Ueno, Akira Naruse, Rio Yokota, and Satoshi
  Matsuoka.
\newblock Large-scale distributed second-order optimization using
  kronecker-factored approximate curvature for deep convolutional neural
  networks.
\newblock In \emph{Proceedings of the IEEE Conference on Computer Vision and
  Pattern Recognition}, pages 12359--12367, 2019.

\bibitem[Pan et~al.(2017)Pan, Innanen, and Liao]{hf2017}
Wenyong Pan, Kristopher~A Innanen, and Wenyuan Liao.
\newblock Accelerating {Hessian}-free {Gauss}-{Newton} full-waveform inversion
  via l-{BFGS} preconditioned conjugate-gradient algorithm.
\newblock \emph{Geophysics}, 82\penalty0 (2):\penalty0 R49--R64, 2017.

\bibitem[Pauloski et~al.(2020)Pauloski, Zhang, Huang, Xu, and Foster]{dis2020}
J.~Gregory Pauloski, Zhao Zhang, Lei Huang, Weijia Xu, and Ian~T. Foster.
\newblock Convolutional neural network training with distributed {K-FAC}.
\newblock \emph{arXiv preprint arXiv:2007.00784v1}, 2020.

\bibitem[Qian(1999)]{sgdm}
Ning Qian.
\newblock On the momentum term in gradient descent learning algorithms.
\newblock \emph{Neural Networks}, 12\penalty0 (1):\penalty0 145--151, 1999.

\bibitem[Robbins and Monro(1951)]{sgd}
Herbert Robbins and Sutton Monro.
\newblock A stochastic approximation method.
\newblock \emph{The Annals of Mathematical Statistics}, pages 400--407, 1951.

\bibitem[Simonyan and Zisserman(2014)]{net2014}
Karen Simonyan and Andrew Zisserman.
\newblock Very deep convolutional networks for large-scale image recognition.
\newblock \emph{arXiv preprint arXiv:1409.1556}, 2014.

\bibitem[Tieleman and Hinton(2012)]{rms2012}
Tijmen Tieleman and Geoffrey Hinton.
\newblock Lecture 6.5-rmsprop: Divide the gradient by a running average of its
  recent magnitude.
\newblock \emph{COURSERA: Neural Networks for Machine Learning}, 4\penalty0
  (2):\penalty0 26--31, 2012.

\bibitem[Yang et~al.(2020)Yang, Xu, Li, Wen, and Chen]{w2020}
Minghan Yang, Dong Xu, Yongfeng Li, Zaiwen Wen, and Mengyun Chen.
\newblock Sketchy empirical natural gradient methods for deep learning.
\newblock \emph{arXiv preprint arXiv: arXiv:2006.05924}, 2020.

\bibitem[Zhang et~al.(2018)Zhang, Sun, Duvenaud, and Grosse]{nkfac2018}
Guodong Zhang, Shengyang Sun, David Duvenaud, and Roger Grosse.
\newblock Noisy natural gradient as variational inference.
\newblock In \emph{International Conference on Machine Learning}, pages
  5847--5856, 2018.

\bibitem[Zhang et~al.(2019)Zhang, Wang, Xu, and Grosse]{wd2019}
Guodong Zhang, Chaoqi Wang, Bowen Xu, and Roger Grosse.
\newblock Three mechanisms of weight decay regularization.
\newblock In \emph{International Conference on Learning Representations}, 2019.

\end{thebibliography}


\end{document}